\documentclass{article}
 \usepackage{graphicx}
 \graphicspath{ {./} }

\usepackage{iclr2024_conference,times}

\usepackage{amsmath,amsfonts,bm}

\def\eqref#1{equation~\ref{#1}}

\def\1{\bm{1}}

\DeclareMathAlphabet{\mathsfit}{\encodingdefault}{\sfdefault}{m}{sl}
\SetMathAlphabet{\mathsfit}{bold}{\encodingdefault}{\sfdefault}{bx}{n}

\usepackage[hidelinks]{hyperref}
\usepackage{url}
\usepackage{xspace}
\usepackage[capitalise]{cleveref}
\usepackage{enumitem}

\crefformat{section}{\S#2#1#3} 
\crefformat{subsection}{\S#2#1#3}
\crefformat{subsubsection}{\S#2#1#3}

\title{GoEX: Perspectives and Designs Towards a 
Runtime for Autonomous LLM Applications}

\newcommand{\eat}[1]{}

\newcommand{\goex}{GoEx\xspace}
\newcommand{\goexlong}{Gorilla Execution Engine\xspace}

\iclrfinalcopy

\begin{document}
\maketitle

\begin{center}
\vspace{0.2in}
    \textbf{
    Shishir G. Patil \quad Tianjun Zhang \quad  Vivian Fang \quad  Noppapon C. \quad  Roy Huang \quad  Aaron Hao} \\
    \vspace{0.1in}
    \textbf{Martin Casado}$^1$ \textbf{\quad  Joseph E. Gonzalez \quad  Raluca Ada Popa \quad  Ion Stoica}
\vspace{0.1in}

    UC Berkeley \quad $^1$Andreessen Horowitz

\vspace{0.1in}

    \texttt{shishirpatil@berkeley.edu}
    
\end{center}

\vspace{0.2in}

\begin{abstract}

Large Language Models (LLMs) are evolving beyond their classical role of providing information within dialogue systems to actively engaging with tools and performing actions on real-world applications and services. 
Today, humans verify the correctness and appropriateness of the LLM-generated outputs (e.g., code, functions, or actions) before putting them into real-world execution. 
This poses significant challenges as code comprehension is well known to be notoriously difficult. 
In this paper, we study how humans can efficiently collaborate with, delegate to, and supervise autonomous LLMs in the future.
We argue that in many cases, ``post-facto validation''---verifying the correctness of a proposed action after seeing the output---is much easier than the aforementioned ``pre-facto validation'' setting.
The core concept behind enabling a post-facto validation system is the integration of an intuitive \emph{undo} feature, and establishing a \emph{damage confinement} 
for the LLM-generated actions as effective strategies to mitigate the associated risks. 
Using this, a human can now either revert the effect of an LLM-generated output or be confident that the potential risk is bounded. 
We believe this is critical to unlock the potential for LLM agents to interact with applications and services with limited (post-facto) human involvement. 
We describe the design and implementation of our open-source runtime for executing LLM actions, \goexlong (\goex), and present open research questions towards realizing the goal of LLMs and applications interacting with each other with minimal human supervision. 
We release \goex{} at \url{https://github.com/ShishirPatil/gorilla/}.

\end{abstract}

\section{introduction}
\label{sec:intro}

Large Language Models (LLMs) are evolving from serving knowledge passively in chatbots to actively interacting with applications and services. Enabling agents and software systems to interact with one another has given rise to new innovative applications. In fact, it is no longer science fiction to imagine that many of the interactions on the internet are going to be between LLM-powered systems. Agentic systems~\citep{wang2023voyager, wu2023autogen, park2023generative}, co-pilots~\citep{roziere2023code}, plugins~\citep{ChatGPTplugins}, function calling and tool use~\citep{patil2023gorilla,qin2023toolllm,parisi2022talm, yao2022react,openai2023gpt4}, are all steps towards this direction. 

\emph{The logical next-step in this evolution is towards autonomous LLM-powered microservices, services, and applications. 
This paper is a first step towards realizing this goal, and addresses some of the key trustworthiness challenges associated with it.
}


As a running example, consider an LLM-powered personal assistant that has access to a user's email account. 
The user asks the assistant to send an important email to their boss, but instead, the LLM sends a sensitive email to the wrong recipient.
In designing such a system, several critical challenges must be addressed:

\textbf{Hallucination, stochasticity, and unpredictability.}
LLM-based applications place an unpredictable and hallucination-prone LLM at the helm of a system traditionally reliant on trust.
Currently, services and APIs assume a human-in-the-loop, or clear specifications to govern how and which tools are used in an application. In our running example, the user clicks the ``Send'' button after confirming the recipient and body of the email.
In contrast, an LLM-powered assistant may send an email that goes against the user's intentions, and may even perform actions unintended by the user.
LLMs are not only capable of being stochastic, but crucially, are capable of unpredictable and unbounded behavior even when trained not to do so~\citep{anilmany}.

\textbf{Unreliability.}
Given their unpredictability and impossibility to comprehensively test, it is difficult for a user to trust an LLM off the shelf.
Consequently, LLM-powered applications are challenging for users to adopt as LLMs are untrusted components that would be running within a trusted execution context.
Trivially, one can ensure safety by restricting the LLM to have no credentials, at the expense of losing utility.
Given that prior work \citep{openai2023gpt4, patil2023gorilla, schick2023toolformer} has surfaced the growing utility of LLM-based systems, mechanisms are needed to express the safety-utility tradeoff to developers and users.

\textbf{Delayed feedback and downstream visibility.}
Lastly, from a system-design principle, unlike chatbots and agents of today, LLM-powered systems of the future will not have immediate human feedback. This means that the intermediate state of the system is not immediately visible to the user and often only downstream effects are visible.
An LLM-powered assistant may interact with many other tools (e.g., querying a database, browsing the web, or filtering push notifications) before composing and sending an email. Such interactions before the email is sent are invisible to the user.

\subsection{A Runtime that Enables Autonomous LLMs}
The question this paper tries to address is: How do we enable an untrusted agent to take sensitive actions (e.g., code generation, API calls, and tool use) on a user's behalf and then verify that those actions aligned with the intent of that user's request?

Traditional methods such as using containers---which guarantee isolation through virtualization---falter when adjustments to the environment are required within the user's context. For example, a user might want to modify the state of their operating system. Further, isolation alone cannot  ensure the final state aligns with the \texttt{intended} actions, especially under ambiguity or execution errors and ambiguity of stated intent. Both are common flaws when intent is being specified in natural-language as opposed to a more precise domain-specific language. 

\begin{figure}
    \centering
    \includegraphics[width=\columnwidth]{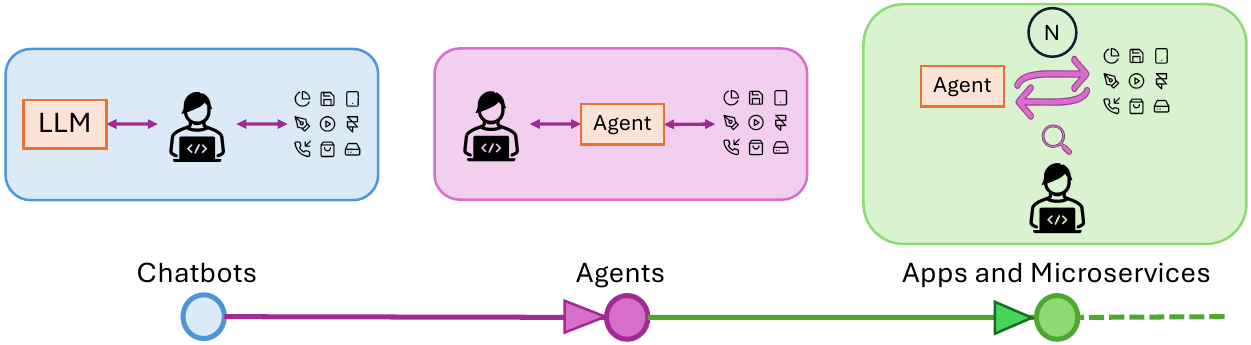}
    \caption{Evolution of LLMs powered applications and services from chatbots, to decision-making agents that can interact with applications and services with human-supervision, to autonomous LLM-agents interacting with LLM-powered apps and services with minimal and punctuated human supervision.
    }
    \label{fig:evolution}
\end{figure}

In this paper, we introduce the notion of \textbf{``post-facto LLM validation''} as opposed to ``pre-facto LLM validation''.
While in both scenarios humans are the ultimate arbitrators, in ``post-facto validation'' human's arbitrate the output produced by executing the actions produced by the LLM, as opposed to the process or the intermediate outputs. 
A natural risk arising from ``post-facto validation'' is that the actions we execute may have unintended consequences. While the benefits of evaluating the output might justify the risks involved with unintended execution, we recognize that this could be a non-starter in many applications. To remedy this, we introduce the intuitive abstractions of ``undo'' and ``damage confinement'' or ``damage confinement''. 
The ``undo'' abstraction allows LLMs to back-track an action that may be unintended, for example, delete a message that was sent in Slack. And for those actions which may not have an ``undo'', we present ``damage confinement'' semantic. ``Damage confinement'' can be considered as a quantification of the user's risk appetite. For example, a user may tolerate the risk of the LLM agent delivering pizza to the wrong address, but perhaps they might not want to allow their LLM to interact with their bank.

As a step towards realizing our vision, we developed \goex{}, a runtime for executing actions generated by LLMs. By designing \goex{} to exclusively utilize readily available off-the-shelf software components, we aim to assess the readiness of current resources and provide an ecosystem to empower developers.

\newpage

In summary, we make the following contributions:

\begin{enumerate}[leftmargin=*]
    \item We first make the case for the future of deeply-embedded LLM systems with LLMs powering microservices, applications, etc. In this paradigm, LLMs are not merely used for information compression, but also as decision makers~(\cref{sec:background}).
    We identify the key challenges associated with integrating LLMs into existing systems, including the inherent unpredictability of LLMs, the lack of trust in their execution, and the difficulty in detecting and mitigating their failures in real-time~(\cref{sec:challenges}).
    \item We introduce the concept of ``post-facto LLM validation'' as an approach to ensuring the safety and reliability of LLM-powered systems, focusing on validating the results of LLM-generated actions rather than the process itself~(\cref{ss:post-facto}). 
    \item We introduce ``undo''~(\cref{ss:reversibility}) and ``damage confinement''~(\cref{ss:blast-radius}) abstractions as mechanisms for mitigating the risk of unintended actions taken in LLM-powered systems.
    \item We propose the {\em \goexlong (\goex)}, a runtime designed to enable the autonomous interactions of LLM-powered software systems and agents by safely executing LLM-generated actions and striking a tradeoff between safety and utility~(\cref{sec:sys-design}).
\end{enumerate}

\section{Evolution of LLM powered Agents}
\label{sec:background}
We first present the  background on the evolution of LLM-powered-systems to mean applications, microservices, and other systems that integrate- or interface- with LLMs. Based on this trend, we then speculate on what such a future would look like.

\subsection{LLM-Human Interaction: Chatbots, and Search}

LLMs have transformed the landscape of human-computer interaction. With early adoption as chatbots, these models were designed to mimic human conversation, allowing users to interact with computers in natural language. This era of LLMs focused on understanding and generating answers, serving as a bridge for humans to interact with vast amounts of web data in a more intuitive way. Early implementations were primarily used in the \emph{read-only} model for information retrieval where the LLM did not make any stateful changes~\citep{burns2022discovering}, customer service agents in enterprises, and as educational tools, laying the foundation for more sophisticated applications.

\subsection{From Chatbots to Agents: The Rise of Actionable LLMs}

Increasing trustworthiness~\citep{wang2023decodingtrust}, the availability of adapters, and novel-techniques~\citep{schick2023toolformer, patil2023gorilla}, have expanded the role of LLMs from passive providers of information to active agents capable of executing simple tasks. These agents, powered by LLMs, can interact with applications, services, and APIs to perform actions on behalf of the user. This shift represents a significant leap in the capabilities of LLMs, enabling them to contribute actively to workflows and processes across various domains.

However, this evolution also brings challenges. The complexity of understanding context, intent, and the subtleties of human language make it difficult to ensure the accuracy and appropriateness of the actions taken by LLMs. As a result, \emph{human oversight} remains crucial to manage and validate the set of actions proposed by the LLM.

\subsection{Towards Ubiquitous LLM Integration}

Looking ahead, we expect the integration of LLMs into daily workflows and systems to deepen. This future envisions LLMs not just as tools or assistants but as pervasive agents embedded in a myriad of workflows, enhancing functionality and adaptability across the board. This vision is materialized through the development of advanced LLM-powered microservices, LLM-powered applications, and LLM-powered workflows all interacting with each other, constantly, with limited-to-no human interaction. 

From an application developer's perspective, these can be categorized into personalized systems, hosted agents for collective use, and third-party integrations.

\paragraph{Personalized LLM-powered workflows for individuals.}

In the personal domain, LLMs are anticipated to become deeply integrated with individual user experiences, offering tailored assistance that understands and anticipates the unique preferences and needs of each user. Imagine a personalized LLM-powered version of Siri or Google Voice, not merely responding to queries but proactively managing schedules, filtering information based on user preferences, and even performing tasks across a range of applications and services. Such personalized systems would mark a significant departure from generic assistants to truly personalized digital companions.

\paragraph{Hosted agents for enterprise and group applications.}

Within professional and enterprise environments, hosted LLM-powered agents will take on specific roles, such as managing database queries or automating routine administrative tasks, tailored to the unique needs of an organization. These specialized agents would be operating within the confines of an organization but would serve a wide set of users.

\paragraph{Third-party agents: expanding the ecosystem of services.}

The expansion of LLM capabilities is also expected to include extensive collaboration with third-party service providers (e.g., Slack, Gmail, Dropbox, etc.), enabling seamless interactions between users and services through personalized LLM workflows. These third-party agents would allow users to communicate with and through them using customized LLM-powered workflows, and integrate a wide range of services.

\section{Navigating the New Frontier: Challenges in Ubiquitous LLM Deployments}
\label{sec:challenges}

With advancements in LLM capabilities and their applications come new challenges:
How do we contend with the inaccuracies inherent to even the best models currently available? How do we ensure security for client information? How do we handle system reliability and quantify risk tolerance to a user?
In the following sections, we elaborate on these new challenges.

\subsection{Delayed Signals}

\textbf{The challenge of timely feedback.} In traditional software development, immediate feedback through error messages or direct outputs enables quick recovery mechanisms to be triggered. However, when embedding LLMs, especially in complex systems or applications interfacing with real-world data and actions, feedback can be significantly delayed. This delay in obtaining relevant signals to assess the performance or correctness of LLM actions (especially with text-in, text-out modality) introduces challenges in rapidly iterating and refining model outputs.

\textbf{Impact on system development.} The lag between action and feedback complicates the identification of errors and the assessment of system performance, potentially leading to additional state being built on top of the system. This necessitates designing systems that can accommodate these delays and implement strategies for asynchronous feedback collection.

\subsection{Aggregate Signals}

In the realm of LLMs, particularly when applied to large-scale systems or microservices, individual LLM actions may not provide clear insights into overall system performance, nor assist in diagnosing the cause of the error. Instead, the true measure of success (or failure) often emerges from aggregated outcomes, necessitating a shift in how developers and stakeholders evaluate and account for, in LLM-driven applications.

\subsection{The Death of Unit-Testing and Integration-Testing}

The integration of LLMs into software systems challenges traditional paradigms of unit testing and integration testing. While closed-source and continuous-pre-trained LLMs from third-parties pose a new challenge of the model changing constantly, in-house LLMs are not a panacea either. Given the dynamic, often unpredictable nature of LLM outputs, establishing a fixed suite of tests that accurately predict and verify all potential behaviors becomes increasingly difficult, if not impossible. 

\subsection{Variable Latency}

LLMs' auto-regressive text-generation by the very nature means inference time may vary as the LLM can either output a long or short response. This is an important consideration for hard-deadline real-time systems (RTS)~\citep{quigley2009ros}.

\subsection{Protecting Sensitive Data}
\label{ss:protecting-sensitive-data}
In order for LLMs to interact with a user's accounts across multiple applications, the LLM must be able to reason about having access to credentials granting access to the user's accounts. When an LLM is hosted by an (untrusted) external service, it is desirable to not directly pass any credentials or sensitive data to the LLM while still preserving functionality of the LLM-powered system.
LLMs can also generate (untrusted) code and are susceptible to hallucinations~\citep{rawte2023surveyhallucination,zhang2023siren}, which can result in running potentially malicious code or inadvertently performing actions that were not intended by the user.

\section{Designing a Runtime}

To address the challenges introduced by the new paradigm of LLM-powered applications, we propose an \emph{LLM runtime}---designed specifically to execute actions generated by LLMs---as a compelling solution. This section discusses the necessity of such a runtime, its envisioned properties, and strategies to mitigate the risks associated with LLM-generated actions.

\paragraph{LLMs started the problem, can LLMs solve the problem?} Despite the rapid advancements in LLMs, expecting LLMs to self-correct and eliminate all potential errors or unintended actions through existing training techniques—--such as pre-training, instruction tuning, DPO~\citep{rafailov2024direct}, or RLHF~\citep{ziegler2019fine}---is promising (\cref{sec:sys-design}) but challenging. The challenges are manifold, primarily due to ill-defined metrics and the inherent complexities of accurately predicting the real-world impacts of actions suggested by LLMs. Thus, while LLMs are at the heart of these challenges, their current evolutionary trajectory suggests they cannot entirely solve the problem without external frameworks to guide their execution.

\subsection{Post-facto LLM validation}
\label{ss:post-facto}
In the realm of LLM-powered-systems, we introduce ``post-facto LLM validation,'' which contrasts with traditional ``pre-facto'' methods. In ``post-facto validation,'' humans evaluate the outcomes of actions executed by the LLM, rather than overseeing the intermediate processes. This approach assumes that validating results over processes, acknowledging that while verifying outcomes is crucial, understanding and correcting processes based on those outcomes is equally important.

Forgoing ``pre-facto validation'' means execution of actions without prior validation, which introduces risks and potentially leads to undesirable outcomes. We propose two abstractions to mitigate the risk associated with post-facto validation: undoing an action~(\cref{ss:reversibility}), and damage  confinement~(\cref{ss:blast-radius}).

\subsubsection{Reversibility}
\label{ss:reversibility}
When possible, actions executed by an LLM should give users the right to \textit{undo} an action. 
This approach may require maintaining multiple versions of the system state, leading to high costs in terms of memory and computational resources. Furthermore, the feasibility of implementing undoing an action is often dependent on the level of access granted to the system. For instance, in file systems or databases, where root access is available, undoing actions is possible. However, in scenarios where such privileged access is not granted, such as in email clients like Gmail, the ability to undo an action may be limited or require alternative approaches. 
One potential solution is for the runtime to make a local copy of the email before deleting, which introduces additional state to the runtime but enables \emph{undo} for email deletion.

To account for the resource costs associated with maintaining multiple system states, we adopt the notion of a \textit{commit}, also called a ``watermark'' in streaming data-flow systems~\citep{carbone2015apache,akidau2015dataflow}. By grouping together sets of actions based on their associativity, commutativity, and distributive properties, it may be possible to define checkpoints at which the system state can be saved or rolled back. This approach would enable selective undoing of actions within a defined scope, rather than maintaining the ability to undo every individual action.

\paragraph{Atomiticy.} In agent-driven systems, the option for users to demand atomicity of operations can be crucial. Atomicity ensures that either all of the operations within a task are successfully completed, or, in the event of a failure in any step, the system is reverted to its initial state before any operation was applied. This binary outcome---success or a clean-slate reset---provides a clear, predictable framework for managing tasks, increasing the system's reliability and user trust in the LLM agent executing complex sequences of actions.

\subsubsection{Damage Confinement}
\label{ss:blast-radius}
Not all applications or tools provide the ability to undo an action. For example, emails currently cannot be unsent after some time has elapsed. In scenarios like these, we fall back to ``damage confinement'' or ``blast-radius confinement'', as it is necessary to provide users with mechanisms to quantify and assess the associated risks of the actions their LLM-powered application may take.

One approach to address this challenge is through the implementation of coarse-grained access control mechanisms. A user could permit their LLM to only read emails instead of sending emails, thus confining the blast radius to an tolerable level.
Such permissioning has already been explored preliminarily by~\citet{wu2024secgpt}, in the context of a user authorizing independent LLM applications to interact with one another.

\subsection{Symbolic Credentials and Sandboxing}

As mentioned in~\cref{ss:protecting-sensitive-data}, the LLM could be (1) hosted by an (untrusted) external provider and (2) susceptible to hallucinations~\citep{rawte2023surveyhallucination,zhang2023siren}, resulting in code that may be unsafe to execute.

In order to protect sensitive user information from and untrusted LLM, the sensitive information in the input prompt can be substituted with a symbolic credential (e.g., a dummy API key), similar to the anonymization approach taken by Presidio~\citep{presidio}, and sending this sanitized prompt to the LLM. Then, the LLM will never see the user's sensitive information in its input.

We can mitigate this risk of running potentially unsafe code by executing the generated code in a sandboxed environment, whether it be a container or a bare-metal VM. With this approach, we only grant the code access to the required dependencies and necessities, such as a specific API key used for service access, and nothing more. If utilizing a container, we only mount the necessary files and impose appropriate network restrictions.

\subsection{Storing Keys and Access Control}
As LLMs are inherently untrusted, a user would feel uneasy permitting the LLM based system to store their credentials.
There are two challenges an LLM runtime should address:
how to store and manage the user's credentials; and determining and projecting the minimal set of permissions that an LLM needs in order to accomplish its task.
More formally, mapping an action to the least privileges that need to be granted to perform the task. While solutions such as asking a ML model have the benefits of generalizability and scalability, pre-computing this permission set manually provides strong security guarantees. Finding a common ground between these two techniques, remains an interesting open area for research. Further,
in an enterprise use case in which an LLM-powered application is managing many user credentials, recording an audit trail of credential access is critical.

\section{GoEx: LLM Runtime}
\label{sec:sys-design}

\goex{} represents a first step towards building a runtime for executing LLM-generated actions within a secure and flexible runtime environment.
Central to \goex{} are its abstractions for ``undo''~(\cref{ss:reversibility}) and ``damage confinement'' or ``blast-radius confinement''~(\cref{ss:blast-radius}), which provide developers of apps and services the flexibility to tailor policies to their specific needs, recognizing the impracticality of a one-size-fits-all policy given the varied contexts in which LLMs are deployed. 
\goex{} supports a range of ``actions'' including  RESTful API requests~(\cref{sec:sys-design-restful}), databases operation~(\cref{sec:sys-design-db}), and filesystem actions~(\cref{sec:sys-design-fs}). Each action type, while initiated from a unified \goex{} interface, are handled uniquely as described below.

\subsection{RESTful API calls}
\label{sec:sys-design-restful}

\begin{figure}[h!]
    \centering
    \includegraphics[width=\columnwidth]{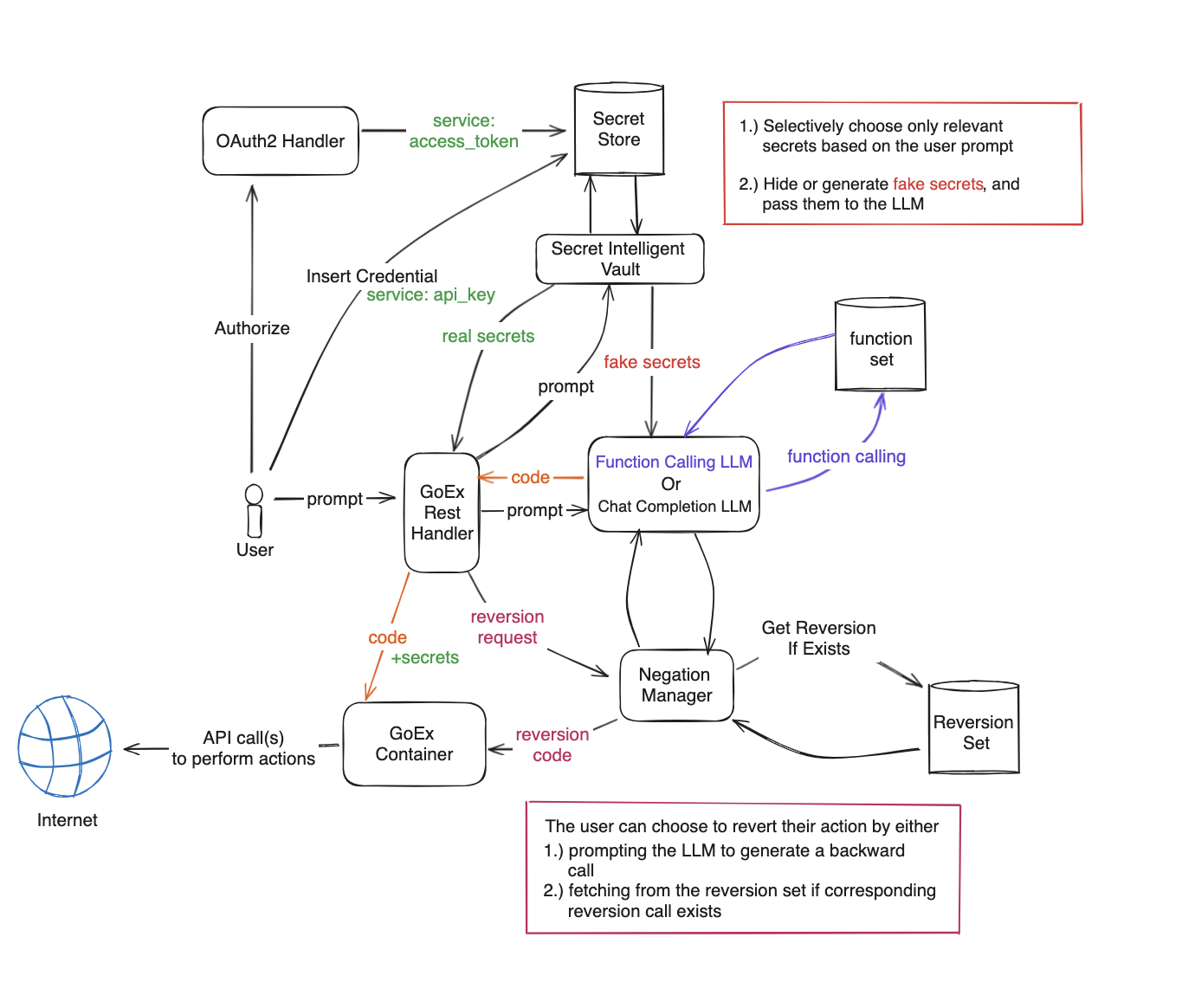}
    \caption{
    \goex{}'s runtime for executing RESTful API calls. Upon receiving the user's prompt, \goex{} presents two alternatives. First, an LLM can be prompted to come up with the (\texttt{Action}, \texttt{Undo-Action}) pair. Second, the application developer can provide tuples of actions and their corresponding undo-actions (function calls) from which the LLM can pick amongst.
    }
    \label{fig:goex-restful}
\end{figure}

We first describe how \goex{} handles RESTful API calls (illustrated in~\cref{fig:goex-restful}).

\paragraph{Authentication.} \goex{} provides a secure way to handle user secrets, whether using OAuth2 for token-based authentication or API keys for direct service access. \goex{} acts as the secure intermediary to facilitate authenticated actions across various services. For OAuth2, \goex{} sits between the user and services, facilitating the necessary relay to retrieve access tokens. These tokens allow users to delegate the \goex{} system to perform actions on their behalf. For other services that authenticate accounts through API keys, \goex{} provides an interface that allows users to insert and retrieve them.

\paragraph{Storing secrets.} User secrets and keys are stored locally on the user's device in a Secret Intelligent Vault (SIV). SIV maps \texttt{service\_name} to \texttt{key} and \texttt{format}. When user wishes to interact with specific service(s), the corresponding keys are requested from the SIV. The \texttt{format} specifies how the keys are store, that is, in a file, or as a string, etc. The role of the SIV is to selectively retrieve just the required keys for a given execution. For example, if a user wants to send an email invite to their friend for lunch, the agent only needs their OAuth2 token for their email provider, and not, for example, their bank account's API keys. The policy used for SIV is user-defined and highly flexible; it could be as simple as parsing through the user prompt to detect which service's keywords are present, or as complex as a fine-tuned prompt-to-service retrieval model. 

\eat{
Despite this, users still face the risk of exposing their credentials to the LLM provider if it becomes malicious. SIV combats this by utilizing the concept of ``dummy secrets'' and ``references.''

With dummy secrets, \goex{} replaces the user's secret with a dummy value of the same format and length. Unlike symbolic execution (e.g., replacing API keys with \texttt{api\_key}), using dummy secrets can help the LLM better understand the datatype of the object they are interacting with and we empirically find it to perform better. On the other hand, the concept of a ``reference'' refers to storing secrets in files, and SIV only passes the paths to those files to the LLM, instructing the generated code to read the key from the file. Typically, OAuth2 works better using a cover because access tokens are commonly stored as a file format.

Once we retrieve the command from the LLM, we can either replace the dummy secret with the real secret or load the secret pointed to by the reference at runtime, ensuring that sensitive information is not leaked to the LLM provider.
}

\paragraph{Generating actions.} The \goex{} framework pupports two techniques to generate the APIs. In the Chat Completion case, assuming the user prompt is, ``send a Slack message to \texttt{gorilla@yahoo.com},'' the user must initially authorize \goex{} to use their access token through the Slack browser. After receiving the user prompt, \goex{} requests the SIV for the necessary secrets from the Secret Store. Slack secrets (OAuth2) are inherently hidden because they are stored as a file, so \goex{} passs the file path along with the prompt directly to the LLM. \goex{} mounts the Slack secret file and passes the LLM-generated code to be executed in the GoEx container. If the user wishes to revert the execution, the reversion call will be retrieved from the reversion set if it exists; otherwise, the handler prompts the LLM to generate it.
If the user chooses Function Calling, instead of asking the LLM to come up with a command to satisfy the user's prompt, \goex{} asks it to select a function from a user-defined function set and populate the arguments. Secrets will be chosen from the SIV similarly, and execution occurs in the GoEx container. If the user wishes to revert, another function from the function set will be chosen by the LLM.

\paragraph{Generating undo actions.} Identifying the `undo' action for RESTful APIs, includes the following steps. First, we check if the reverse call for the action API is in the database \texttt{Reversion Set} as shown in figure~\ref{fig:goex-restful}. 
\goex{} presents the systems abstractions, while  developers are free to define the policies for mapping. For some APIs it might be critical to check for exact match for all parameters of the API, on the other hand for some other APIs, perhaps just the API name might be sufficient to uniquely identify what the reverse API would be. 
For example it is 
\emph{not} 
sufficient to say the reverse of \texttt{send\_slack\_message} is
\texttt{delete\_slack\_message}, since number of messages to be deleted could be one of the arguments. 

To populate such a mapping, first, we instruct the LLM to generate a reverse API call whenever the user attempts to perform an action. We recognize that this gives no guarantees, but the  philosophy is that we allow the LLM to be wrong at most once. Post each new API, the table is then if the reversion worked or not making this information available for future invocations. 
For applications that need guarantee, developers can pre-populate this table and combined with function-calling mode of operation, the system can be forced to only use those API's that are `guaranteed' by the developers to be reversible. 

\paragraph{Damage confinement.} Often reversibility cannot be guaranteed. For examples sending an email isn't really reversible. For such scenarios, \goex{} presents abstraction to bound the worst case. Currently, the way blast-radius-containment is implemented is through coarse-grained access control, and exact string match. First, \goex{} looks at the user's prompt to determine the end service that they are then authorized to use. For example, a prompt of \emph{I would like to send a slack message} would only need credentials for slack, and not, say, their bank. \goex{} currently does this, through a simple sub-string check of the prompt, while giving developers the flexibility to adopt any mapping they might choose. 

\paragraph{Execution.} Once the API, and the set of credentials required are determined, the APIs are then executed in a Docker container for isolation.

\subsection{Database Operations}
\label{sec:sys-design-db}

\goex{} leverages the mature transaction semantics offered by databases. This section describes the abstractions available, and the two default policies.

\begin{figure}[th!]
    \centering
    \includegraphics[width=\columnwidth]{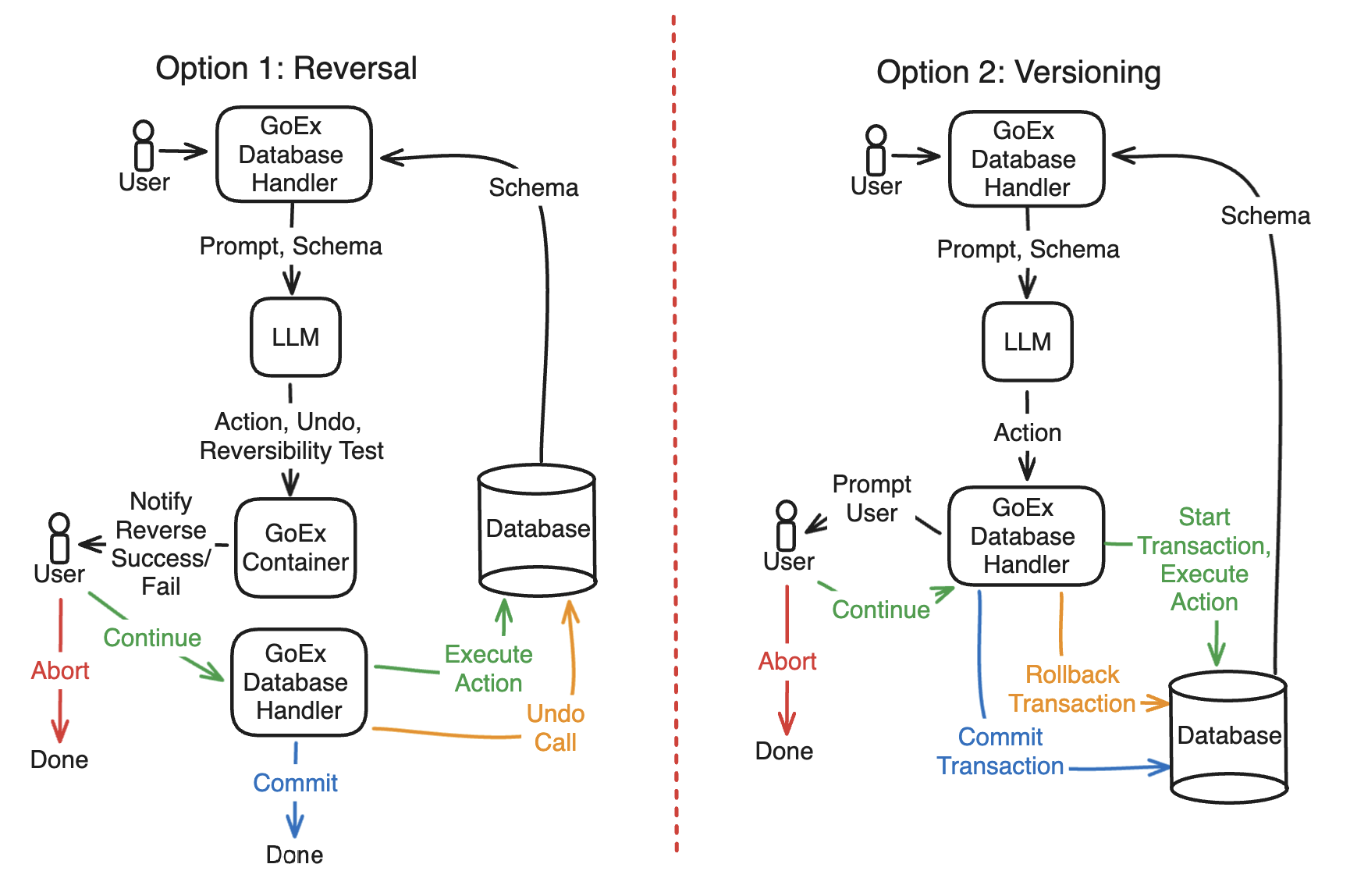}
    \caption{Runtime for executing actions on a database. We present two techniques to determine if a proposed action can be undone.
    On the left, for non-transactional databases like MongoDB, and for flexibility, we prompt the LLM to generate (\texttt{Action}, \texttt{Undo-Action}, \texttt{test-bed}) tuples, which we then evaluate in a isolated container to catch any false (\texttt{Action}, \texttt{Undo-Action}) pairs.
    On the right, we can provide a deterministic undo with guarantees by employing the transaction semantics of databases.
    }
    \label{fig:database}
\end{figure}

\subsubsection{Abstractions}

\goex{} relies on the LLM to generate database operations, but there are two prerequisites needed to execute database operations: (1) knowledge of the current database state, and (2) knowledge on how to access the database. To provide these, \texttt{DBManager} class is used. This allows the database to readily minimally query for the database state (e.g. only the schema) to provide additional info to the LLM during prompting without leaking sensistive data. It also tracks the connection configuration to the database so that connections can be established without leaking credentials to the LLM as an untrusted third-party by asking the user to store the credentials locally, and after the LLM generates the operation, \goex{} then executes the operation.

\texttt{DBManager} also assists the user store with storing a previous state. Here, the \textit{commit} and \textit{undo} actions are introduced where a \textit{commit} means the user permanently saves the executed changes, and an \textit{undo} reverses the aforementioned changes. Most modern databases also provide ACID guarantees~\citep{haerder1983principlesacid}, including NoSQL databases like DynamoDB and MongoDB, which we leverage to implement committing and undoing actions.

\subsubsection{Policy}

\texttt{DBManager} implements reversibility in two ways. The user chooses which one to use when they execute a prompt in \goex{}.

\begin{enumerate}[leftmargin=*]
    \item \textbf{Option 1 (Reversal).} Makes use of a reverse database operation to perform the \textit{undo}. It is done by prompting the LLM with the original operation (action call) along with the schema to generate the reversal operation (undo call). Committing would require no action, and undoing would just be performing the undo call after the action call is done. This option scales better as additional users can continue perform database actions without needing to wait for the previous user to finish their transaction at the cost of relying on the LLM to come up with an undo call, which may or may not have unexpected behaviors.
    
    \item \textbf{Option 2 (Versioning).} Makes use of the traditional ACID transaction guarantees of the database and holds off on completing a transaction until the user specifies to do so, or rolls back to the previous state. Committing would involve committing the transaction, and undoing is synonymous to a rollback transaction. This branch is able to provide reversal guarantees that branch 1 cannot, at the expense of higher performance overhead.
\end{enumerate}

\textbf{Reversibility testing.} Within Option 1, \goex{} also performs a reversibility test to verify that the generated reversal operation indeed reverses the original operation. This requires a containerized environment to be separate from the original database to maintain the original database state. Since copying over the database into the container is very expensive, the approach is to ask the LLM to generate a bare-bones version of the database for reversibility testing, given the action, undo calls, and the database schema. The outcome of the test is sent back to the user for final confirmation before committing or undoing the operation. This method allows for efficient testing by decoupling the testing runtime from being scaled by the number of entries in the database. \

\subsection{File Systems}
\label{sec:sys-design-fs}

\goex{} tries to present expressive abstractions to let LLM-powered systems to interact with file-systems using Git version control. To track the directory tree, on every \goex{} filesystem-type execution, \goex{} does an exhaustive, recursive walk of the directory and its subdirectories and stores the directory structure as a formatted string.

\begin{figure}
    \centering
    \includegraphics[width=\columnwidth]{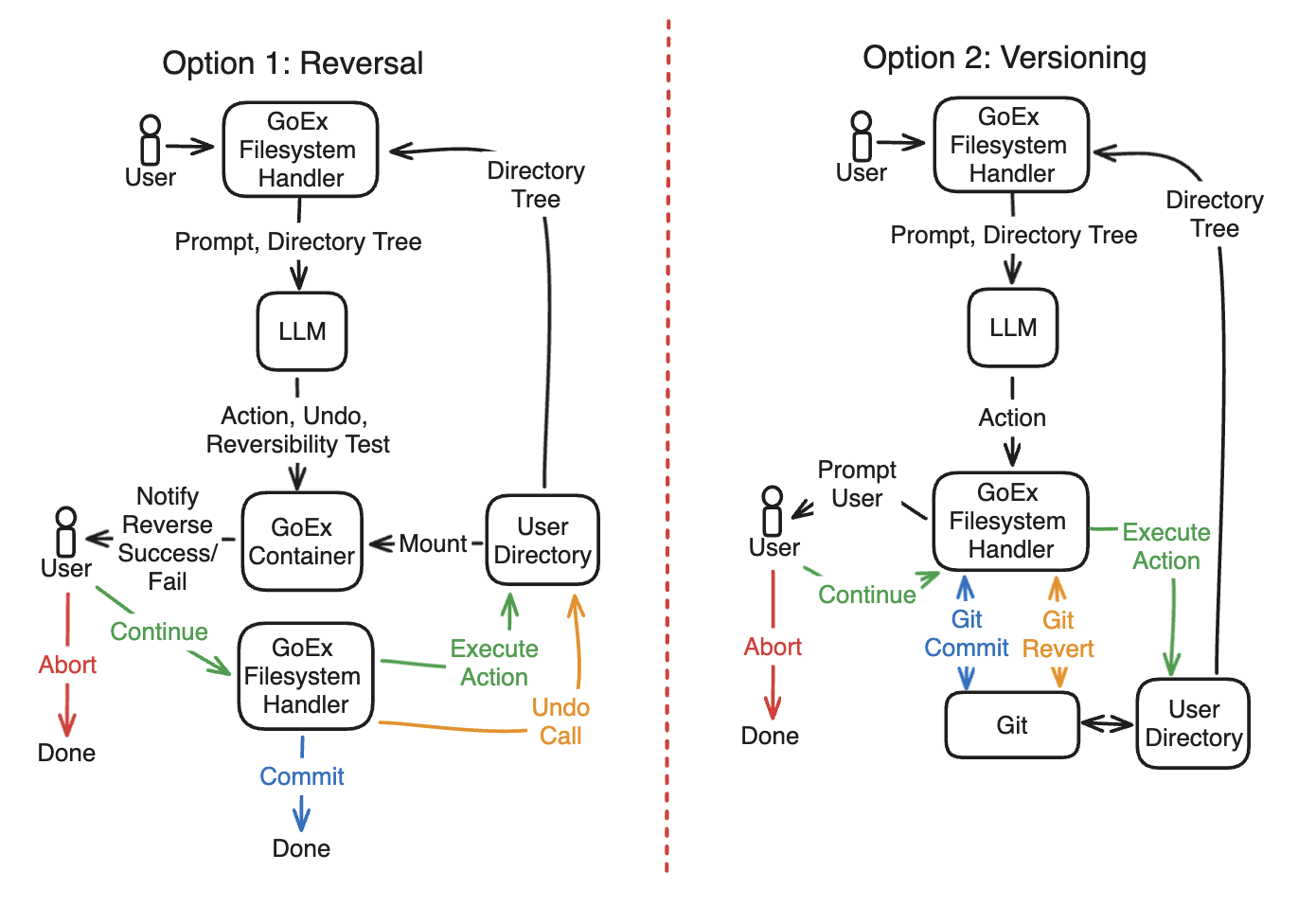}
    \caption{Runtime for executing actions on a filesystem. \goex{} presents two abstractions. On the left, the LLM is prompted to come up with an (\texttt{Action}, \texttt{Undo-Action}, \texttt{test-bed}) which \goex{} evaluates in a isolated container to catch any false (\texttt{Action}, \texttt{Undo-Action}) pairs.
    On the right presents deterministic guarantees by using versioning control system like Git or Git LFS. 
    }
    \label{fig:filesystem}
\end{figure}

\subsubsection{Abstractions}

Filesystems operation support in \goex{} uses abstractions similar to what is used to support database operations.
\texttt{FSManager}, is a filesystems manager that tracks (1) the directory tree structure with all filenames, and (2) the directory path that the user wishes to execute the filesystem's operations in. The tree structure, which is updated with executions, enables the LLM to generate operations that reflect the actual state of the user's filesystem. 

Utilizing the relevant abstractions presented by journaling and log-structured filesystem for undo-semantics is left as future work, as the current \goex{} system aims for compatibility.

\subsubsection{Policy}

The options are similar to the database case, where Option 1 is for reversals and Option 2 is for versioning. The largest differences are how \texttt{FSManager} carries out reversibility testing and that versioning is accomplished using Git.

\paragraph{Git.} \goex{} uses Git to perform versioning. Since Git is already a version-control system for files, it is a straightforward solution to use, but has several limitations. Git does not have the ability to version track outside of the directory that it was initialized in. GoEx limits the user execution scope to the specified path in \texttt{FSManager}---which is always inside of a Git repository---and its subdirectories in accordance to our blast-radius confinement abstraction to prevent the LLM from performing arbitrary actions in undesired parts of the user's system. With larger directories, Git versioning can be expensive space-wise. GoEx leverages Git LFS for larger directories as an optimization. A threshold is defined for directory size that \goex{} would then check whether or not to initialize Git LFS (200 MB by default).

\textbf{Reversibility testing.} Similar to supporting databases operations, the LLM generates the testing code using the action and undo calls, along with the directory tree. Inside the container, the specified path is mounted in read-only mode to again do blast radius containment. 
\goex{} begins by duplicating the directory contents in the container, then run the action and undo calls on the copied directory, and finally compare contents. Depending on the original operation, the content comparison can just be a check of filenames or an exhaustive file content comparison of all the files. We rely on the LLM to come up with the test-case. Unsurprisingly, here \goex{} allows you to trade off guarantees for performance.

\section{Discussion}

\subsection{Is Post-Facto LLM Validation Always Preferable?} 
It is not lost on us that while post-facto LLM validation has many benefits, our advocacy of it is also somewhat philosophical. For example, if one were to bake a cake it's probably better to taste the cake than check the recipe. But on the the contrary, if one were to produce an audit report, it might be preferable to check the process.  We acknowledge both options---post-facto LLM validation, and pre-facto LLM validation---as two techniques to evaluate the LLM's actions, however, this paper focuses on verifying results as this is, perhaps, more appropriate in most scenarios we consider, which are complex microservice settings.

\subsection{Designing LLM-friendly APIs}

The conversation around LLM-powered systems design is predominantly centered around designing systems to conform with the API semantics of existing applications and services. However, an equally interesting question is what API design in an LLM-centric world would looks like.

LLMs introduce a paradigm where applications and services can anticipate and adapt to the intricacies of LLM interactions. A notable feature that embodies this adaptability is the implementation of ``dry-run'' semantics, akin to the functionality commonly visible in infrastructure products such as~\citet{awsdryrun}, ~\citet{kubernetesdryrun}, and where API calls can be tested to predict their success without executing any real changes. This concept can be extended beyond mere prediction, serving as a bridge between LLMs' proposed actions and user consent. By repurposing ``dry-run'' operations, service providers can offer a preview of the uncommitted state resulting from an LLM's actions, allowing users to evaluate and approve these actions before they are finalized. This process adds an essential layer of user oversight, ensuring that actions align with user expectations and intentions.

\paragraph{Chaining-aware.} Applications and Services should by-default expect their APIs to be chained with each other when used by agents. To support such a scenario, there needs to be a way to express which APIs can be commutative, associative or distributive with a given set of APIs.  

\subsection{Tracking LLM agents}
The introduction of a nonce mechanism (i.e., a session identifier) would enable LLMs to present their identity and facilitate smoother interactions with API providers. This could serve various purposes, such as identifying a session initiated by an LLM or providing a context for transactions. A transaction ID, for example, can enable a system to identify and potentially rollback actions based on this ID. This not only improves the traceability of interactions but also contributes to the overall robustness and reliability of the system by providing an auditable framework for LLM-powered systems.

\section{Related Work}
\paragraph{Isolation.} Prior work on enabling automated LLM-powered systems~\citep{wu2024secgpt} draws on concepts from existing computer systems~\citep{cohen1975protection, reis2019site, wilkes1979cambridge, linden1976operating} and emphasizes isolation between LLM-powered applications in order to secure the overall system.
Isolation is one facet of safely executing LLM-powered systems, as it alone cannot ensure that the final execution outcome aligns with the user's intended action.

\paragraph{Trusthworthiness in LLMs.}
There is a rich body of work benchmarking LLMs on their robustness~\citep{chang2023survey}. Recently, trustworthiness~\citep{wang2023decodingtrust} has been introduced as a multifaceted benchmark encapsulating robustness, stereotype bias, toxicity, privacy, ethics, and fairness. \citet{wang2023decodingtrust} found that more advanced LLMs (e.g., GPT-4) exhibit higher, albeit still imperfect, trustworthiness.

\paragraph{Attacks on LLMs and defenses.}
LLMs are also susceptible to prompt hacking attacks, of which include prompt injection~\citep{perez2022ignore,greshake2023not,yu2023assessing,schulhoff2023hackaprompt,liu2023prompt} and jailbreaking~\citep{kang2023exploiting,anilmany}. Such attacks can lead to unpredictable and malicious decisions made by the LLM.
There is also an active line of work on defending against such attacks on LLMs~\citep{piet2023jatmo,chen2024struq,suo2024signed,toyer2023tensor,yi2023benchmarking}. 
These attacks and defenses will continue to evolve, and consequently the potential of LLMs being susceptible to having their trustworthiness undermined necessitates a runtime that can provide execution of LLM-decided actions while limiting risk.

\section{Conclusion}
The evolution of LLMs from chatbots to deeply embedding them in applications and services for autonomous operation among themselves and other agents presents an exciting future. 
In this paper, we introduce the concept of ``post-facto LLM validation,'' as opposed to pre-facto LLM validation, to enable users to verify and roll back the effects caused by executing LLM generated actions (e.g., code, API invocations, and tool use).
We propose \goex{}, a runtime for LLMs with an intuitive undo and damage confinement abstractions, enabling the safer deployment of LLM agents in practice. 
We hope our attempt to formalize our vision and present open research questions towards realizing the goal of autonomous LLM-powered systems in the future, is a step towards a world where LLM-powered systems can independently, with minimal human verification, interact with other tools and services.

\bibliography{iclr2024_conference}
\bibliographystyle{iclr2024_conference}

\end{document}